\documentclass[twocolumn, switch]{article} 

\usepackage{preprint}

\usepackage{amsmath, amsthm, amssymb, amsfonts}
\usepackage{bm}
\usepackage{mathtools}

\usepackage[numbers,square]{natbib}
\usepackage[utf8]{inputenc}	
\usepackage[T1]{fontenc}	
\usepackage{xcolor}		
\usepackage{url}
 
\usepackage{hyperref}
\hypersetup{colorlinks = true,
            linkcolor = blue,
            urlcolor  = blue,
            citecolor = blue,
            anchorcolor = black
}
\usepackage{booktabs} 		
\usepackage{nicefrac}		
\usepackage{microtype}		
\usepackage{lineno}		
\usepackage{float}			

\usepackage{caption}
\usepackage{array}
\usepackage[ruled,vlined]{algorithm2e}

\usepackage{newfloat}
\DeclareFloatingEnvironment[name={Supplementary Figure}]{suppfigure}

\usepackage{titlesec}
\titlespacing\section{0pt}{12pt plus 3pt minus 3pt}{1pt plus 1pt minus 1pt}
\titlespacing\subsection{0pt}{10pt plus 3pt minus 3pt}{1pt plus 1pt minus 1pt}
\titlespacing\subsubsection{0pt}{8pt plus 3pt minus 3pt}{1pt plus 1pt minus 1pt}

\usepackage{graphics}
\usepackage{multirow}
\usepackage{hhline}
\usepackage{subfigure}

\title{Towards Visual Distortion in Black-Box Attacks}

\usepackage{authblk}

\author{Nannan Li}
\author{Zhenzhong Chen\thanks{\tt{zzchen@ieee.org}}}
\affil{School of Remote Sensing and Information Engineering, Wuhan University}

\begin{document}

\twocolumn[ 
  \begin{@twocolumnfalse} 
  
\maketitle

\begin{abstract}
Constructing adversarial examples in a black-box threat model injures the original images by introducing visual distortion. In this paper, we propose a  novel black-box attack approach that can directly minimize the induced distortion by learning the noise distribution of the adversarial example, assuming only loss-oracle access to the black-box network. The quantified visual distortion, which measures the perceptual distance between the adversarial example and the original image, is introduced in our loss whilst the gradient of the corresponding non-differentiable loss function is approximated by sampling noise from the learned noise distribution. We validate the effectiveness of our attack on ImageNet. Our attack results in much lower distortion when compared to the state-of-the-art black-box attacks and achieves $100\%$ success rate on InceptionV3, ResNet50 and VGG16bn. The code is available at \url{https://github.com/Alina-1997/visual-distortion-in-attack}.
\end{abstract}
\vspace{0.35cm}

  \end{@twocolumnfalse} 
] 



\section{Introduction}
Adversarial attack has been a well-recognized threat to existing Deep Neural Network (DNN) based applications. It injects small amount of noise to a sample (e.g., image, speech, language) but degrades the model performance drastically \cite{audio20,KurakinGB17,practical17}. With the continuous improvements of DNN, such attack could cause serious consequences in practical conditions where DNN is used. According to \cite{survey18,towards}, adversarial attack has been a practical concern in real-world problems, ranging from cell-phone camera attack to attacking self-driving cars.

According to the information that an adversary has of the target network, existing attack roughly falls into two categories: white-box attack that knows all the parameters of the target network, and black-box attack that has limited access to the target network. Each category can be further divided into several subcategories depending on the adversarial strength \cite{PapernotLimit16}. The proposed attack in this paper belongs to loss-oracle based black-box attack, where the adversary can obtain the output loss from supplied inputs. In real-world scenario, it's sometimes difficult or even impossible to have full access to certain networks, which makes the black-box attack practical and attract more and more attention. 

\begin{figure*}[!t]
\centering
    \includegraphics[width=0.75\textwidth]{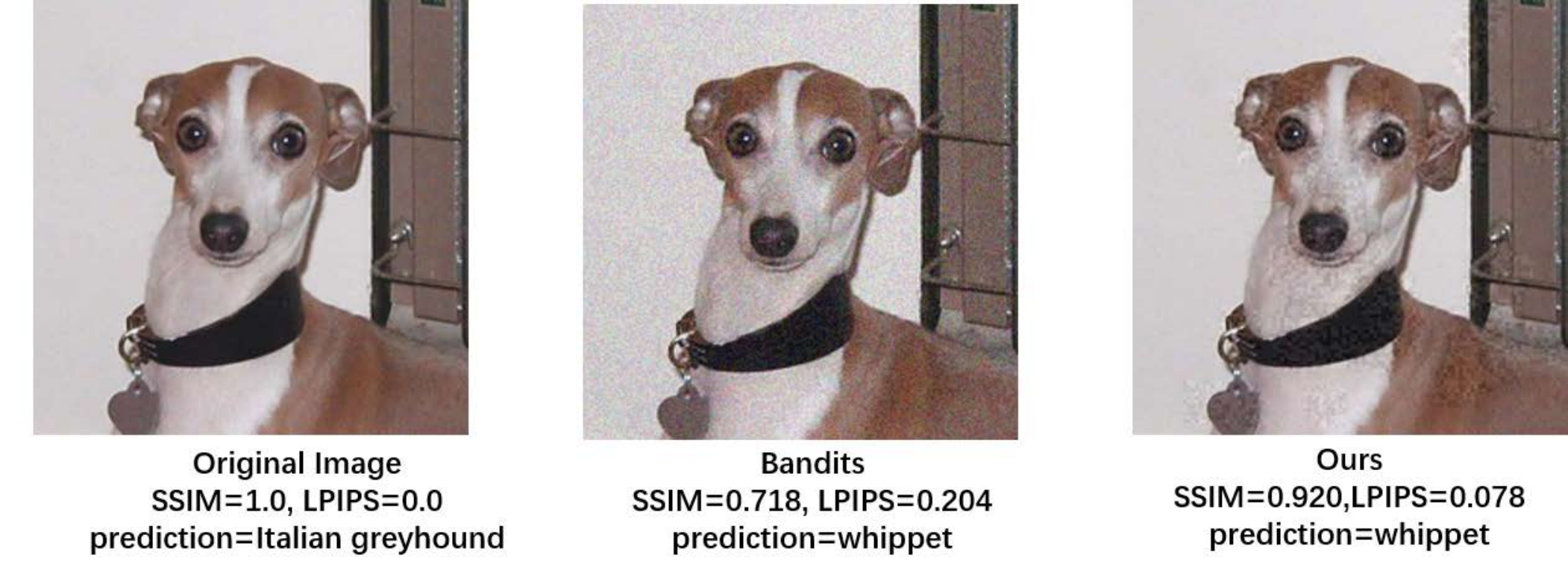}
  \caption{Adversarial examples on ImageNet with bounded noise ${||\delta||}_\infty \leq 0.05$. The first image is the original unperturbed image. The following examples are from \cite{ilyas2018prior} and our method, respectively. Higher Structural SIMilarity (SSIM) and lower Learned Perceptual Image Patch Similarity (LPIPS) indicate less visual distortion.}
    \label{fig:motiv}
\end{figure*}

Black-box attack has very limited or no information of the target network and thus is more challenging to perform. In the $l_p$-bounded setting, a black-box attack is usually evaluated on two aspects: number of queries and success rate. In addition, recent work \cite{jordan2019quantifying} shows that visual distortion in the adversarial examples is also an important criteria in practice. Even under a small $l_\infty$ bound, perturbing pixels in the image without considering the visual impact could make the distorted image very annoying. As shown in Fig. \ref{fig:motiv}, an attack \cite{ilyas2018prior} under a small noise level ($l_\infty \leq 0.05$) causes relatively large visual distortion and the perturbed image is more distinguishable from the original one. Therefore, under the assumption that the visual distortion caused by the noise is related to the spatial distribution of the perturbed pixels, we take a different view from previous work and focus on \emph{explicitly} learning a noise distribution based on its corresponding visual distortion.

In this paper, we propose a novel black-box attack that can directly minimize the induced visual distortion by learning the noise distribution of the adversarial example, assuming only loss-oracle access to the black-box network. The quantified visual distortion, which measures the perceptual distance between the adversarial example and the original image, is introduced in our loss where the gradient of the corresponding non-differentiable loss function is approximated by sampling noise from the learned noise distribution. The proposed attack can achieve a trade-off between visual distortion and query efficiency by introducing the weighted perceptual distance metric in addition to the original loss. Theoretically, we prove the convergence of our model under a convex or non-convex loss function. The experiments demonstrate the effectiveness of our attack on ImageNet. Our attack results in much lower distortion than the other attacks and achieves $100\%$ success rate on InceptionV3, ResNet50 and VGG16bn. In addition, it is shown that our attack is valid even when it's only allowed to perturb pixels that are out of the target object in a given image.

Our contributions are as follows:
\begin{itemize}
\item We are the first to introduce perceptual loss in a \emph{non-differentiable} way for the generation of less-distorted adversarial examples. And the proposed method can achieve a trade-off between visual distortion and query efficiency by using the weighted perceptual distance metric in addition to the original loss. 
\item Theoretically, we prove the convergence of our model.
\item Through extensive experiments, we show that our attack results in much lower distortion than the other attacks.
\end{itemize}

\section{Related Work}
Recent research on adversarial attack \cite{there20, pca20,pso20} has made advanced progress in developing strong and computationally efficient adversaries. In the following, we briefly introduce existing attack techniques in both the white-box and black-box settings. 

\subsection{White-box Attack} 
In white-box attack, the adversary knows the details of a network, including network structure and its parameter values. Goodfellow \emph{et al.} \cite{GoodfellowSS14} proposed a fast gradient sign method to generate adversarial examples. It's computationally effective and serves as a baseline for attacks with additive noise. In \cite{functional19}, a functional adversarial attack that applied functional noise instead of additive noise to the image, is introduced. Recently, Jordan \emph{et al.} \cite{jordan2019quantifying} stressed quantifying perceptual distortion of the adversarial examples by leveraging perceptual metrics to define an adversary. Different from our method which directly optimizes the metric, their model conducts a search over parameters from several composed attacks. There are also attacks that sample noise from a noise distribution \cite{r12zheng2019distributionally,r13wang2020hamiltonian}, on the condition that gradients from the white-box network is accessible. Specifically, \cite{r12zheng2019distributionally} utilizes particle approximation to optimize a convex energy function. \cite{r13wang2020hamiltonian} formulates the attack problem as generating a sequence of adversarial examples in a Hamiltonian Monte Carlo framework. 

In summary, white-box attack is hard to detect or defend \cite{advCarlin}. In the meantime, however, it suffers from label-leaking and gradient-masking problem \cite{KurakinGB17}. The former causes adversarially trained models to perform better on adversarial examples than original images, and the latter neutralizes the useful gradient for adversaries. The preliminary of acquiring full access to a network in white-box attack is also sometimes difficult to satisfy in real-world scenarios.

\subsection{Black-box Attack}
Black-box attack considers the target network as a black-box, and has limited access to the network. We discuss loss-oracle based attack here, where the adversary assumes only  loss-oracle access to the black-box network. 

\paragraph{Query Efficient Attacks.} Attacks of this kind roughly fall into three categories: 1) Methods that estimate gradient of the black-box. Some methods estimate the gradient by sampling around a certain point, which formulates the task as a problem of continuous optimization. Tu \emph{et al.} \cite{TuTC0ZYHC19} searched for perturbations in the latent space of an auto-encoder. \cite{r1zhang2020dual} utilizes feedback knowledge to alter the searching directions for efficient attack. Ilyas \emph{et al.} \cite{ilyas2018prior} exploited prior information about the gradient.  Al-Dujaili and O`Reilly \cite{there20} reduced query complexity by estimating just the sign of the gradient. In \cite{r2linonlinear,r3_5huang2019black}, the proposed methods perform search in a constructed low-dimensional space. \cite{ld2019} shares similarity with our method as it also \emph{explicitly} defines a noise distribution. However, the distribution in \cite{ld2019} is assumed to be an isometric normal distribution without considering visual distortion whilst our method does not assume the distribution to be a specific form. We compare with their method in details in the experiments. Other approaches in this category develop a substitute model \cite{practical17,ChengDPSZ19,papernot2016transferability} to approximate performance of the black-box. By exploiting the transferability of adversarial attack \cite{GoodfellowSS14}, the white-box attack technique applied to the substitute model can be transferred to the black-box. These approaches assume only label-oracle to the targeted network, whereas training of the substitute model requires either access to the training dataset of the black-box, or collection of a new dataset. 2) Methods based on discrete optimization. In \cite{MoonAS19,there20}, an image is divided into regular grids and the attack is performed and refined on each grid. Meunier \emph{et al.} \cite{meunier2019yet} adopted the tiling trick by adding the same noise for small square tiles in the image. 3) Methods that leverage evolutionary strategies or random search \cite{meunier2019yet,andriushchenko2019square}. In \cite{andriushchenko2019square}, the noise value is updated by a square-shaped random search at each query.  Meunier \emph{et al.} \cite{meunier2019yet} developed a set of attacks using evolutionary algorithms using both continuous and discrete optimization.

\paragraph{Attacks that Consider Visual Impact.} Query efficient black-box attacks usually do not consider the visual impact of the induced noise, for which the adversarial example could suffer from significant visual distortion. Similar to our work, there are research that address the perceptual distance between the adversarial examples and the original image. \cite{r10xiao2018generating,r11zhang2019generating} introduce Generative Adversarial Network (GAN) based adversaries, where the gradient of the perceptual distance in the generator is computed through backpropagation. \cite{r6gragnaniello2019perceptual,r7rozsa2016adversarial} also require the adopted perceptual distance metric to be differentiable. Computing the gradients of a complex perceptual metric at each query might be computationally expensive \cite{r18gao2015learning}, and is not possible for some rank-based metrics \cite{r19ma2016no}. Different from these methods, our approach treats the perceptual distance metric as a black-box, saving the efforts of computing its gradients, and minimizing such distance by sampling from a learned noise distribution. On the other hand, \cite{r8zhao2017generating,r9wang2020transferable} present semantic perturbations for adversarial attacks. The produced noise map is semantically meaningful to human, whilst the image content of the adversarial example is distinct from that of the original image. Different from \cite{r8zhao2017generating,r9wang2020transferable} that focus on \emph{semantic} distortion, our method addresses \emph{visual} distortion and aims to generate adversarial examples that are visually indistinguishable from the original image.

\section{Method}
\begin{algorithm}[t]
\DontPrintSemicolon
\LinesNumbered
\caption{Our Algorithm}
\KwIn{image $x$, maximum norm $\epsilon$, proportion $q$ of the resampled noise}
\KwOut{adversarial example $x+\delta$}
Initialize noise distribution $p_{\theta_0}=\text{softmax}({\theta}_0)$ and noise ${{\delta}_0}$ \;
\For{$\mathrm{step}$ $t$ $\mathrm{in}$ $\{1,...,n\}$}{
$T^*= {\text{argmin}_{T = 0,1,...t - 1}}L(x,x + {\delta _T})$ \;
Compute baseline $b = {L}(x,x + {\delta}_{T^*} )$ \;
Update $\theta$ using Eq. (\ref{eq:detaE}), ${{\theta}_t} \leftarrow {{\theta}_{t-1}}- {\nabla}{F}(\theta _{t-1})$\;
Sample ${\delta}_t$, ${\delta _{t}} \leftarrow {\text{resample}}{({\delta _{T^*}},q;{\delta _{t-1}})_{{\delta _{t-1}}\sim{p_{{\theta _{t-1}}}}}}$ \;
\If{$successful{\text{\_}}attack(x,x+{\delta_t})$}
{return $x+\delta_t$\;}
}
\SetKwProg{Def}{def}{:}{}
\Def{$successful{\text{\_}}attack(x,x+{\delta_t})$}{
\eIf{${\text{argmax}_{k_1}}{f(x+{\delta _t})_{k_1}}{\neq}{\text{argmax}_{k_2}}{f(x)_{k_2}}$}
     {return True \;}
      {return False \;}
      }
\end{algorithm}
\subsection{Learning Noise Distribution Based on Visual Distortion}
An attack model is an adversary that constructs adversarial examples against certain networks. Let $f:x \to f(x)$ be the target network that accepts an input ${x} \in {\mathbb{R}^n}$ and produces an output ${f(x)} \in {\mathbb{R}^m}$. $f(x)$ is a vector and ${f(x)}_k$ represents its $k_\text{th}$ entry, denoting the score of the $k_\text{th}$ class. $y={ \text{argmax}_k} {{f(x)}_k}$ is the predicted class. Given a valid input ${x}$ and the corresponding predicted class $y$, an adversarial example \cite{SzegedyZSBEGF13} $x^\prime$ is similar to $x$ yet results in an incorrect prediction ${ \text{argmax}_k} {{f(x^\prime)}_k}{\neq}y$. In an additive attack, an adversarial example $x^\prime$ is a perturbed input with additive noise $\delta$ such that $x^\prime=x+{\delta}$.  The problem of generating an adversarial example is equivalent to produce noise map $\delta$ that causes wrong prediction for the perturbed input. Thus a successful attack is to find ${\delta}$ such that ${\text{argmax}_k}{f(x+{\delta})_k}{\neq}y$. Since this constraint is highly non-linear, the loss function is usually rephrased in a different form \cite{towards}:
\begin{equation}
L(x,x + \delta )={\text{max}}(0,f{(x + \delta)_y} - {\text{ma}}{{\text{x}}_{k \ne y}}f{(x + \delta )_k})
\end{equation}
The attack is successful when $L=0$. It's noted that such a loss does not take the visual impact into consideration, for which the adversarial example could suffer from significant visual distortion. In order to constrain the visual distortion caused by the difference between $x$ and $x+\delta$, we adopt a perceptual distance metric $d(x,x + \delta)$ into the loss function with a predefined hyperparameter $\lambda$:
\begin{equation}
\begin{aligned}
L(x,x + \delta ) =&{\text{max}}\big{(}0,f{(x + \delta)_y} - {\text{ma}}{{\text{x}}_{k \ne y}}f{(x + \delta )_k}\big{)} \\
&+ {\lambda}d(x,x + \delta )
\end{aligned}
\label{eq:L}
\end{equation}
\begin{figure*}[!t]
\centering
    \includegraphics[width=0.9\textwidth]{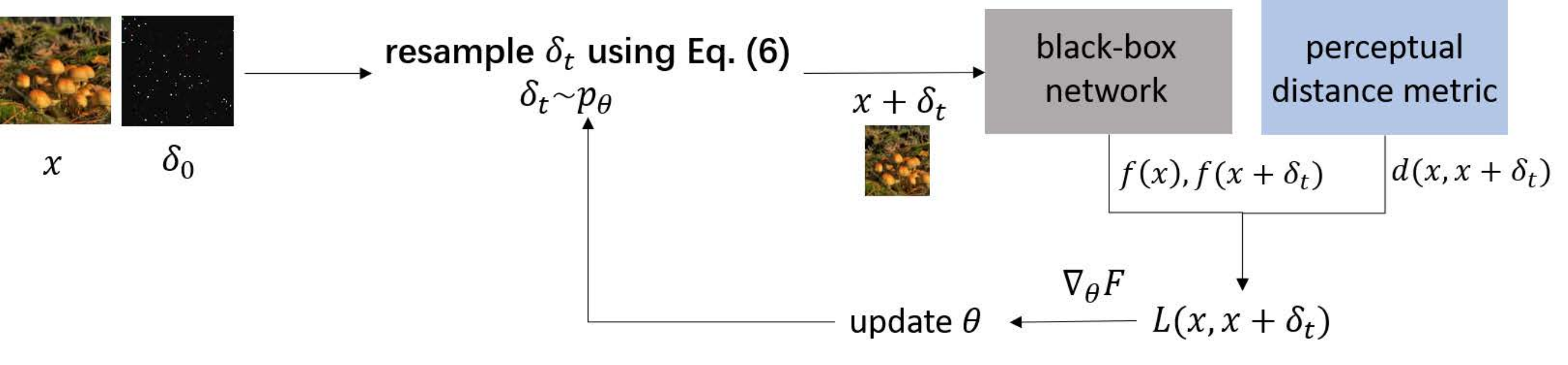}
    \caption{Framework of the proposed attack.}
    \label{fig:model}
\end{figure*}
where smaller $d(x,x + \delta )$ indicates less visual distortion. $d$ can be any form of metric that measures the perceptual distance between $x$ and $x+\delta$, such as well-established $1-\text{SSIM}$ \cite{ssim} or LPIPS \cite{ZhangIESW18}. $\lambda$ manages the trade-off between a successful attack and the visual distortion caused by the attack. The effects of $\lambda$ will be further discussed in Section \ref{sec:abl}.

Minimizing the above loss function faces a challenge that $L$ is not differentiable since the black-box adversary does not have access to the gradients of $L$ and the metric $d(x,x+\delta)$ might be calculated in a non-differentiable way. To address this problem, we \emph{explicitly} assume a flexible noise distribution of $\delta$ in the discrete space, in the sense that the noise values and their probabilities are discrete. And the gradient of $L$ can be estimated by sampling from this distribution. Suppose that $\delta$ follows a distribution $p_{\theta}$ parameterized by $\theta$, \emph{i.e.}, $\delta  \sim {p_\theta}$. For the $j_\text{th}$ pixel in an image, we make its noise distribution $p_{\theta^j}=\text{softmax}(\theta^j)$, where $\theta^j$ is a vector and each element in it denotes a probability value. By sampling noise from the distribution $p_{\theta}$, $\theta$ can be learned to minimize the expectation of the above loss such that the attack is successful (\emph{i.e.}, alters the predicted label) and the produced adversarial example is less distorted (\emph{i.e.}, small $d$):
\begin{equation}
  {\text{minimize}}{\text{ }}\mathbb{E}_{\delta  \sim {p_{\theta}}}[L(x,x + \delta )] 
\end{equation}
For the $j_\text{th}$ pixel, we define its noise's sample space to be a set of discrete values ranging from $-\epsilon$ to $\epsilon$: ${\delta^j}  \in \{ \epsilon ,\epsilon  - \frac{\epsilon }{N},\epsilon  - 2\frac{\epsilon }{N},...,0,... - \epsilon \}$, where $N$ is the sampling frequency and $\frac{\epsilon }{N}$ is the sampling interval. The noise value ${\delta}^j$ of the $j_\text{th}$ pixel is sampled from this sample space by following $p_{\theta^j}$, $p_{\theta^j} \in {\mathbb{R}}^{2N+1}$.

Given $W$ and $H$ the width and height of an image, respectively, since each pixel has its own noise distribution $p_{\theta^j}$ of length $2N+1$, the number of parameters for the entire image is $(2N+1)WH$. Note that we do not consider the difference of color channels in order to reduce the size of the sample space. Otherwise the number of parameters would be tripled. Thus, the same noise value is sampled for each RGB channel of a pixel. To estimate $\theta$, we adopt policy gradient \cite{sutton1998reinforcement} to make the above expectation differentiable with respect to $\theta$. Using REINFORCE, we have the differentiable loss function $F(\theta)$:
\begin{equation}
\begin{aligned}
  F(\theta)&={\mathbb{E}_{\delta  \sim {p_\theta }}}[L(x,x + \delta ) - b] \\
  &= (L(x,x + \delta ) - b)\log ({p_\theta}(\delta))
  \end{aligned}
 \label{eq:E}
\end{equation}
\begin{equation}
\begin{aligned}
  {\nabla}{F}(\theta)&={\nabla _\theta }{\mathbb{E}_{\delta  \sim {p_\theta }}}[L(x,x + \delta ) - b]\\
  & = (L(x,x + \delta ) - b)(1 - {p_\theta}(\delta))
    \end{aligned}
   \label{eq:detaE}
\end{equation}
where $b$ is introduced as a \emph{baseline} in the expectation with specific meaning: 1) when $L(x,x + \delta )<b$, the sampled noise map $\delta$ returns low $L$, and its probability ${p_\theta}(\delta)$ increases through gradient descent; 2) when $L(x,x + \delta )=b$, ${\nabla}{F}(\theta)=0$ and ${p_\theta}(\delta)$ remains unchanged; 3) when $L(x,x + \delta)>b$,  the sampled noise map $\delta$ returns high $L$, and its probability ${p_\theta}(\delta)$ decreases through gradient descent. To sum up,  $L(x,x + \delta)$ is forced to improve over $b$.  At the iteration $t$, we choose $b ={{\text{min}}}_{T = 0,1,...t - 1} {L}(x,x + {\delta}_T )$ such that $L$ improves over the obtained minimal loss.

The above expectation is estimated using a single Monte Carlo sampling at each iteration, and the sampling of noise map $\delta$ is critical. Simply sampling ${\delta}_t$ at the iteration $t$ on the entire image might cause large variance on the norm of the noise, \emph{i.e.}, $||{\delta}_t-{\delta}_{t-1}||_2$. Therefore, to ensure a small variance, with $T^*= {{\text{argmin}}}_{T = 0,1,...t - 1} {L}(x,x + {\delta}_T )$, only a number of $qWH$ pixels' noise values are resampled in $\delta _{T^*}$, while $(1-q)WH$ pixels' noise values remain unchanged:
\begin{equation}
{\delta _{t+1}} \leftarrow {\text{resample}}{({\delta _{T^*}},q;{\delta _t})_{{\delta _t}\sim{p_{{\theta _t}}}}}
\label{eq:resample}
\end{equation}
The above equation means replacing $qWH$ pixels' noise values in noise map $\delta _{T^*}$ with those in ${\delta _t}$, which are sampled from distribution $p_{\theta_t}$. In other words, if $q=0.01$, then only a random $1\%$ of $\delta_{T^*}$ is updated at each iteration.  As shown in Fig. \ref{fig:model}, after sampling ${\delta}_t$, the feedback $L(x,x+{\delta}_t)$ from the black-box and the perceptual distance metric decide the update of the distribution $p_{\theta_t}$. The iteration stops when the attack is successful, \emph{i.e.},  ${\text{max}}(0,f{(x + \delta_t)_y} - {\text{ma}}{{\text{x}}_{k \ne y}}f{(x + \delta_t)_k})=0$.
\subsection{Proof of  Convergence}
Ruan \emph{et al.} \cite{RuanHK18} shows that feed-forward DNNs (Deep Neural Networks) are Lipschitz continuous with a Lipschitz constant $K$, for which we have
\begin{equation}
\forall {t},||f(x + {\delta _t}) - f(x + {\text{ }}{\delta _{{T^*}}})|{|}_2 \leq K||{\delta _t} - {\delta _{{T^*}}}|{|_2}
\label{eq:ineq1}
\end{equation}
At the iteration $t$, since only a small part of the noise map is updated, it can be assumed that
\begin{equation}
|{\text{ma}}{{\text{x}}_{k \ne y}}f{(x + {\delta _t})_k} - {\text{ma}}{{\text{x}}_{k \ne y}}f{(x + {\delta _{{T^*}}})_k}| \leq C
\label{eq:maxineq}
\end{equation}
where $C$ is a constant. Suppose that the perceptual distance metric $d$ is normalized to $[0,1]$. Substituting the inequalities (\ref{eq:ineq1}) and (\ref{eq:maxineq}) in our definition of $L$ in Eq. (\ref{eq:L}) gets the following inequalities:
\begin{equation}
\begin{aligned}
&|L(x,x + {\delta _t}) - L(x,x + {\delta _{{T^*}}})| \\
& \leq K||{\delta _t} - {\delta _{{T^*}}}|{|_2} + C + \lambda \\
& \leq 2KWH\epsilon cq +C+\lambda
 \end{aligned}
\label{eq:12}
\end{equation}

Ideally, $L(x,x+\delta_t)-L(x,x+\delta_{T^*})$ accurately quantifies the difference of the perturbed image even when only one noise value for just a single pixel at the iteration $t$ is different from that at $T^*$. Let $\delta^{ij}$ represent a special noise map, whose $j_\text{th}$ pixel's noise value is the $i_\text{th}$ element in its sample space and the other pixels' noise values are $0$. Note that the length of the sample space for each pixel is $2N+1$. Similarly, ${p_{\theta_t}}(\delta^{ij})$ denotes the probability of the $i_\text{th}$ element in the sample space of the $j_\text{th}$ pixel. By sampling every element in the sample space of the $j_\text{th}$ pixel, we define $l_t^j$ and $p_{\theta _t^j}$ to be a vector:
\begin{gather}
\forall {j \in \{1,2,...,WH\}}, l _t^j = vector[L(x,x + \delta^{ij}) - L(x,x + {\delta _{{T^*}}})], \notag \\
 i = 1,2,...,2N + 1
\end{gather}
 \begin{gather}
 \forall {j \in \{1,2,...,WH\}}, {p_{\theta _t^j} } = vector[{p_{\theta_t}}({\delta}^{ij})], \notag \\
 i = 1,2,...,2N + 1
 \end{gather}

Although the above equations are only meaningful under the ideal situation where $L$ can quantify the difference of just one perturbed pixel, we use these equations for a theoretical proof of convergence. In the ideal situation, the gradient of the $j_\text{th}$ pixel's parameters can be calculated exactly as
\begin{equation}
\nabla F(\theta _t^j) = {l_t^j} \cdot ({\mathbf{1}} - p_{\theta _t^j})
\end{equation}

According to Eq. (\ref{eq:12}) when the number of the resampled pixels $qWH$=1, we have
\begin{equation}
|L(x,x + \delta ^{ij}) - L(x,x + {\delta _{{T^*}}})| \leq 2K{\epsilon}c+C+\lambda
\label{eq:16}
\end{equation}

Note that for $\forall {t_1},{t_2}$ that share the same $T^*$, $l_{t_1}^j$ is equal to $l_{t_2}^j$. Thus, using Eq. (\ref{eq:16}), we have
\begin{equation}
\begin{aligned}
&||\nabla F({\theta _{{t_1}}^j}) - \nabla F({\theta _{{t_2}}^j})|{|_2}\\
&  {\leq} (2N + 1) (2K{\epsilon}c+C+\lambda)||{\text{softmax}}({\theta _{{t_1}}^j}) - {\text{softmax}}({\theta _{{t_2}}^j})|{|_2}
\label{eq:detaE}
\end{aligned}
\end{equation}
In practice, we adopt a single Monte Carlo sampling instead of sampling every noise values for every pixel, for which $2N+1$ should be replaced by $1$ in the above inequality. The inequality (\ref{eq:detaE}) thus becomes:
\begin{equation}
\begin{aligned}
& {||}\nabla F({\theta _{{t_1}}^j}) - \nabla F({\theta _{{t_2}}^j})|{|_2}\\
 & \leq (2K{\epsilon}c + C + \lambda )||{\text{softmax}}({\theta _{{t_1}}^j}) - {\text{softmax}}({\theta _{{t_2}}^j})|{|_2} \\
 &\leq (2K{\epsilon}c + C + \lambda )||{\theta _{{t_1}}^j} - {\theta _{{t_2}}^j}|{|_2}
 \end{aligned}
\end{equation}

The standard softmax function disappears because it is Lipschitz continuous with the Lipschitz constant being $1$ \cite{gao2017properties}. Finally, we have the inequality for ${||}\nabla F({\theta _{{t_1}}}) - \nabla F({\theta _{{t_2}}})|{|_2}$:
\begin{equation}
{||}\nabla F({\theta _{{t_1}}}) - \nabla F({\theta _{{t_2}}})|{|_2} \leq (2K{\epsilon}c + C + \lambda )||{\theta _{{t_1}}} - {\theta _{{t_2}}}|{|_2}
\label{eq:detaE-lip}
\end{equation}

The above inequality proves that $F(\theta)$ is $L$-smooth with the Lipschitz constant being $2K\epsilon c+ C + \lambda$. If $F(\theta)$ is convex, the exact number of steps that Stochastic Gradient Descent (SGD) takes to convergence is $\frac{{(2K\epsilon c+ C + \lambda)\cdot||{\theta _0} - {\theta ^*}||_2^2}}{\xi}$ , where $\xi$ is an arbitrary small tolerable error ($\xi>0$). However, since deep network $L$ is usually highly non-convex, we need to consider the situation where$F(\theta)$ is non-convex. 

Let the SGD update be
\begin{equation}
\theta_{t+1}={\theta_t}+{\eta_t}{g({\theta_t})}
\end{equation}
where ${\eta_t}$ is the learning rate and ${g({\theta_t})}$ is the stochastic gradient. We assume that the variance of the stochastic gradient is upper bounded by $\sigma^2$:
\begin{equation}
\mathbb{E}[||{\nabla}{F(\theta)}-g(\theta)|{|_2^2}] \leq \sigma^2 < \infty
\end{equation}
\begin{table*}[!t]
		\centering
		\caption{Ablation results of the perceptual distance metric, {\upshape $\lambda$} and sampling frequency {\upshape $N$}. Smaller {\upshape $1-\text{SSIM}$}, LPIPS and CIEDE2000 indicate less visual distortion.}
		\begin{tabular}{cccccccc}\toprule
      Sampling &Perceptual &\multirow{2}{*}{$\lambda$}  &Success &\multirow{2}{*}{$1-\text{SSIM}$} &\multirow{2}{*}{LPIPS} &\multirow{2}{*}{CIEDE2000} &Avg. \cr
      Frequency &Metric &&Rate &&&&Queries \cr\midrule
       \multirow{7}{*}{$N=1$}
      &-  &0 &\textbf{100\%} &0.091 &0.099 &0.941 &\textbf{356} \cr  \cmidrule(r){2-8}
       &\multirow{4}{*}{$1-\text{SSIM}$}
        &10 &\textbf{100\%} &0.076 &0.081 &0.741 &401   \cr
         &&100 &97.4\% &0.036 &0.051 &0.703 &1395\cr
         &&200 &92.2\% &0.025 &0.040 &0.622 &2534\cr
         &&dynamic &\textbf{100\%} &\textbf{0.009} &0.009 &\textbf{0.204} &7678\cr \cmidrule(r){2-8}
         &\multirow{4}{*}{LPIPS}
        &10  &\textbf{100\%} &0.080 &0.078 &0.762 &450 \cr
         &&100  &98.1\% &0.049 &0.052 &0.711 &1174\cr
         &&200 &95.1\% &0.038  &0.045 &0.635 &1928 \cr
          &&dynamic &\textbf{100\%} &0.015 &\textbf{0.005} &0.277 &6694 \cr\midrule
         None &$1-\text{SSIM}$ &10 &\textbf{100\%} &0.118 &0.142 &5.936 &426\cr \midrule
        $N=2$ &$1-\text{SSIM}$ &10  &99.7\% &0.071 &0.074 & 0.846 &520 \cr \midrule
         $N=5$  &$1-\text{SSIM}$ &10  &99.5\% &0.069 &0.070 &0.877 &665\cr \midrule
        $N=10$ &$1-\text{SSIM}$ &10  &98.7\% &0.062 &0.075 &0.879 &669 \cr \midrule
       $N=12$ &$1-\text{SSIM}$ &10 &98.7\% &0.071 &0.075 &0.882 &673\cr
		\bottomrule
		\end{tabular}			
		\label{tab:abl}		
\end{table*}
And we select ${\eta_t}$ that satisfies
\begin{equation}
{\sum\limits_{t = 1}^\infty  {{\eta _t}}  = \infty} \text{ and }{\sum\limits_{t = 1}^\infty  {{\eta _t}^2}  < \infty }
\label{eq:condeta}
\end{equation}
Condition (\ref{eq:condeta}) can be easily satisfied with a decaying learning rate, e.g., ${\eta _t} = \frac{1}{{\sqrt {t\ln (t + 1)} }}$. According to Lemma 1 and Theorem 2 in \cite{r21non-convWeb}, using the $L$-smooth property of $F(\theta)$ , ${\nabla}{F}(\theta_t)$  goes to $0$ with probability $1$. This means that with probability $1$ for any $\xi>0$ there exists $N_{\xi}$ such that ${\nabla}{F}(\theta_t){\leq}{\xi}$ for $t{\geq}{N_{\xi}}$. Unfortunately, unlike in the convex case, we do not know the exact number of steps that SGD takes to convergence.

The above proof simply aims to theoretically show that the proposed method converges in finite steps, although possibly in a rather slow speed. From the ``Avg. Queries'' in the following experiments, we can see that the actual computational cost is affordable and comparable to some of the query-efficient attacks. 

\begin{figure*}[!t]
\centering
    \includegraphics[width=0.8\textwidth]{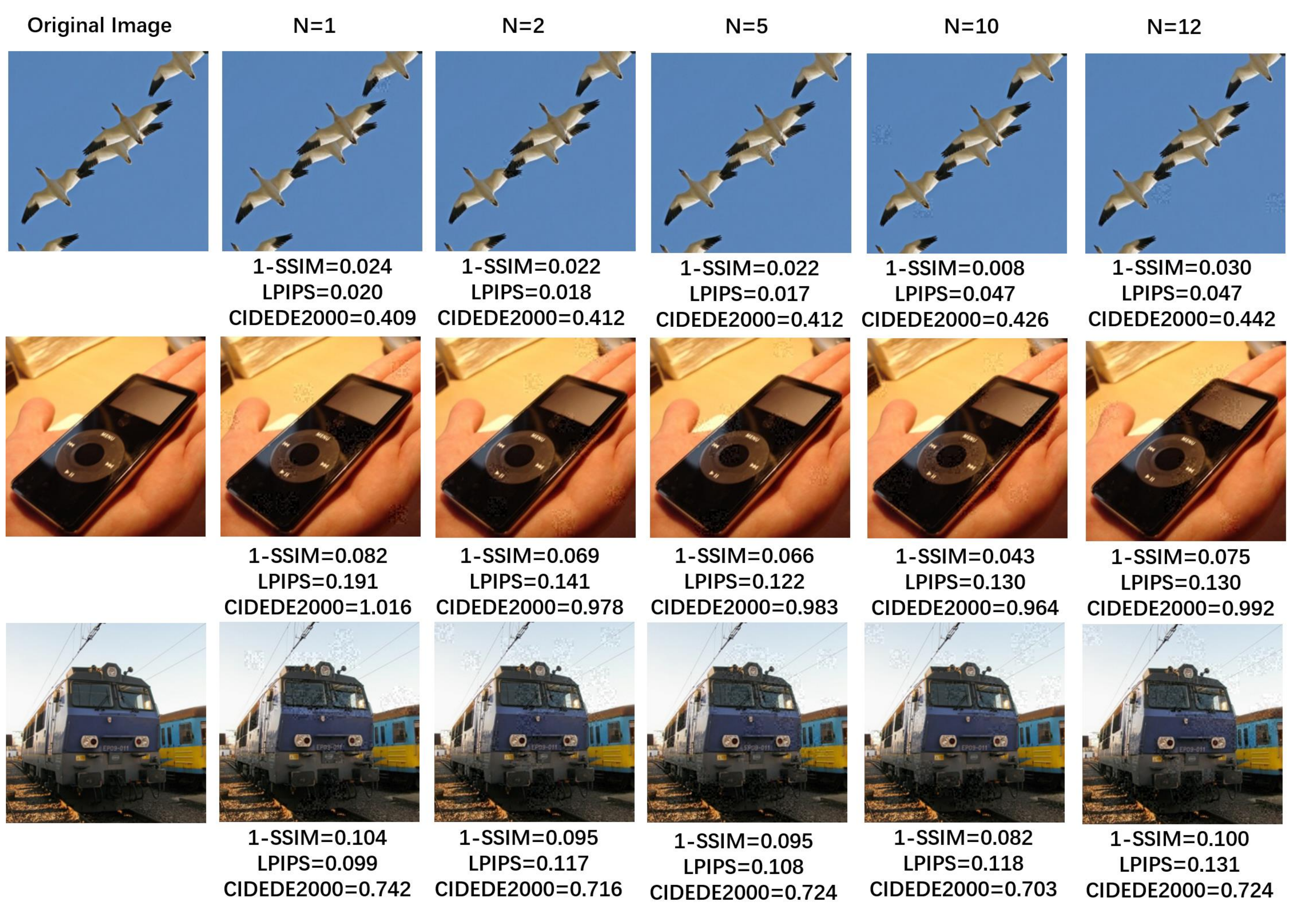}
  \caption{Adversarial examples under different sampling frequency. From left to right is the original image, the adversarial examples from $N=1,2,5,10,12$, respectively.}
  \label{fig:vis_samp}
\end{figure*}
\section{Experiments}
\label{sec:experiments}
Following previous work \cite{meunier2019yet,ilyas2018prior}, we validate the effectiveness of our model on the large-scale ImageNet \cite{ILSVRC15} dataset. We use three pretrained classification networks on Pytorch as the black-box networks: InceptionV3 \cite{incep}, ResNet50 \cite{He2015Deep} and VGG16bn \cite{SimonyanZ14a}. The attack is performed on images that were correctly classified by the pretrained network. We randomly select $1000$ images in the validation set for test, and all images are normalized to $[0,1]$. We quantify our success in terms of the perceptual distance ($1-\text{SSIM}$, LPIPS and CIEDE2000) as we address the visual distortion caused by the attack. In these metrics, $1-\text{SSIM}$ \cite{ssim} measures the degradation of structural information in the adversarial examples. LPIPS \cite{ZhangIESW18} evaluates the perceptual similarity of two images with their normalized distance between their deep features. CIEDE2000 \cite{r20zhao2020towards} measures perceptual color distance, which is developed by the CIE (International Commission on Illumination). Smaller value of these metrics denotes less visual distortion. Except for $1-\text{SSIM}$, LPIPS and CIEDE2000, the success rate and average number of queries are also reported as in most previous work. The average number of queries refers to the average number of requests to the output of the black-box network.

We initialize the noise distribution $p_\theta$ to be a uniform distribution and noise $\delta _0$ to be $0$. The learning rate is $0.01$ and $q$ is set to be $0.01$. In addition, we specify the shape of the resampled noise at each iteration to be a square \cite{meunier2019yet,MoonAS19, andriushchenko2019square}, and adopt the tiling trick \cite{ilyas2018prior,meunier2019yet} with tile size$=2$. The upper bound $\epsilon$ of our attack is set to be $0.05$ as in previous work.
\subsection{Ablation Studies}
\label{sec:abl}
\begin{figure*}[!t]
\centering
    \includegraphics[width=0.9\textwidth]{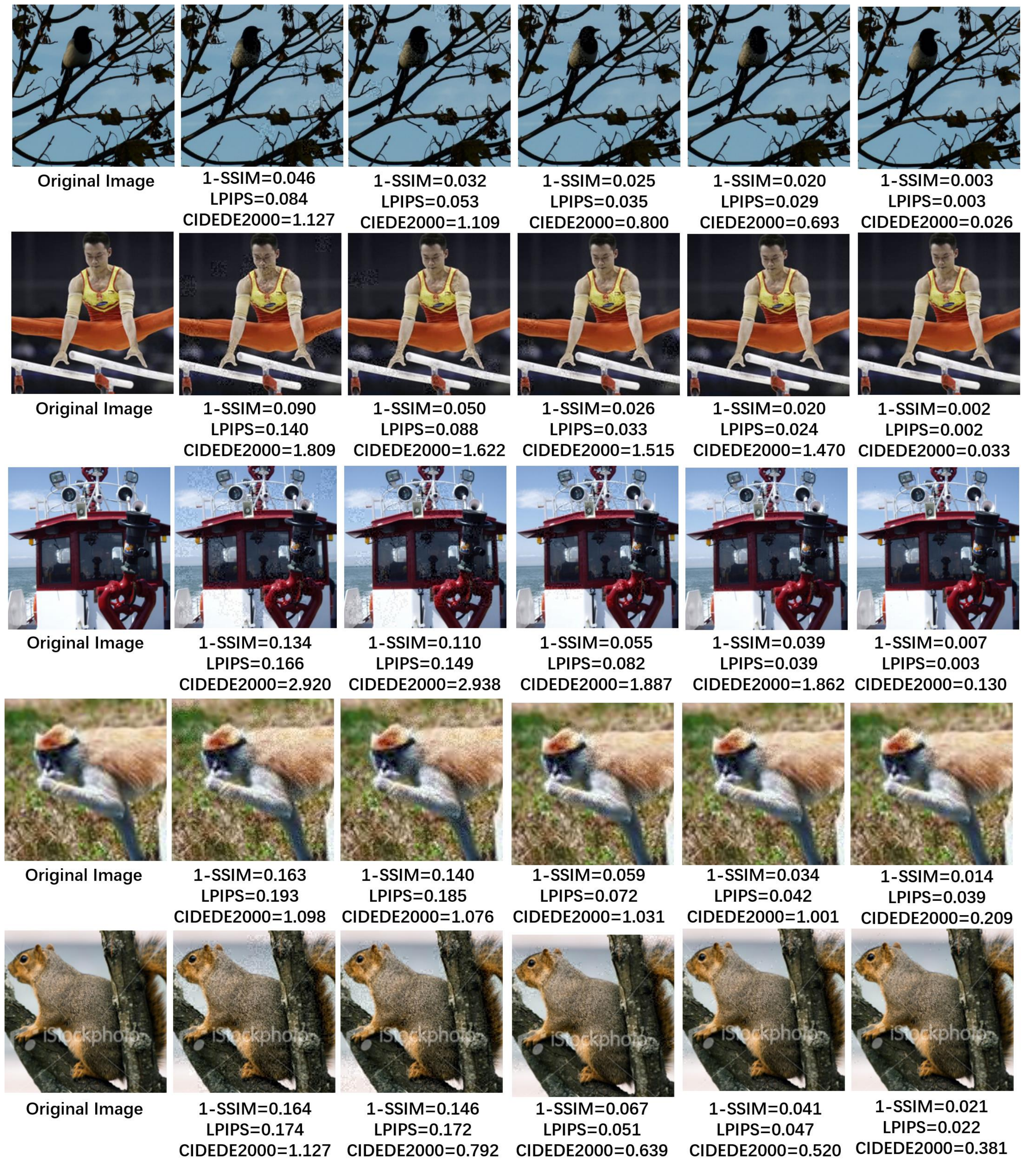}
  \caption{Visualized examples of the proposed attack. From left to right is the original image, the adversarial examples on $\lambda=0, \lambda=10, \lambda=100,\lambda=200$, dynamic $\lambda$, respectively.}
  \label{fig:vis_exp}
\end{figure*}
In the ablation studies, the maximum number of queries is set to be $10,000$. The results are averaged on $1000$ test images. In the following, we discuss the trade-off between visual distortion and query efficiency, the effects of using different perceptual distance metrics in the loss function, the results on different sampling frequencies and the influence of predefining a specific form of noise distribution.
\paragraph{Trade-off between visual distortion and query efficiency.} Under the same $l_\infty$ ball, a query-efficient way to produce an adversarial example is to perturb most pixels with the maximum noise values $ \pm\epsilon$ \cite{MoonAS19,andriushchenko2019square}. However, such attack introduces large visual distortion, which could make the distorted image very annoying. To constrain the visual distortion, the perturbed pixels should be those who cause smaller visual difference while performing a valid attack, which takes extra queries to find. This brings the trade-off between visual distortion and query efficiency, which can be controlled by $\lambda$ in our loss function. As shown in Table \ref{tab:abl}, when $N=1$ and $\lambda=0$, the adversary does not consider visual distortion at all, and perturbs each pixel that is helpful for misclassification until the attack is successful. Thus, it causes the largest perceptual distance ($0.091$, $0.099$ and $0.941$) with the least number of queries ($356$). As $\lambda$ increases to $200$, all the perceptual metrics decrease at the cost of more queries and lower success rate. The maximum $\lambda$ in Table \ref{tab:abl} is $200$ since further increasing it causes the success rate to be lower than $90\%$. In addition, as in \cite{TuTC0ZYHC19}, we perform a dynamic line search on the choice of $\lambda$ to see the best perceptual scores the adversary can achieve, where ${\lambda} \in [0,1000]$. Comparing with fixed $\lambda$ values, using dynamic values of $\lambda$ greatly boosts the performance on perceptual metrics with $100\%$ attack success rate, at the cost of dozens of times the number of queries. Fig. \ref{fig:vis_exp} gives several visualized examples on different $\lambda$, where adversarial examples with larger $\lambda$ suffer from less visual distortion.
%
\paragraph{Ablation studies on the perceptual distance metric.} The perceptual distance metric $d$ in the loss function is predefined to measure the visual distortion between the adversarial example and the original image. We adopt $1-\text{SSIM}$ and LPIPS as the perceptual distance metric to optimize, respectively, and report their results in Table \ref{tab:abl}. When $\lambda=10$, optimizing $1-\text{SSIM}$ shows better score on $1-\text{SSIM}$ ($0.076$ v.s. $0.080$) and CIEDE2000 ($0.721$ v.s. $0.742$) whilst optimizing LPIPS has better performance on LPIPS ($0.078$ v.s. $0.081$). However, when $\lambda$ increases to $100$ and $200$, optimizing $1-\text{SSIM}$ gives better scores on both $1-\text{SSIM}$ and LPIPS. Therefore, we set the perceptual distance metric to be $1-\text{SSIM}$ in the following experiments.
\begin{figure*}[!t]
\centering
    \includegraphics[width=0.9\textwidth]{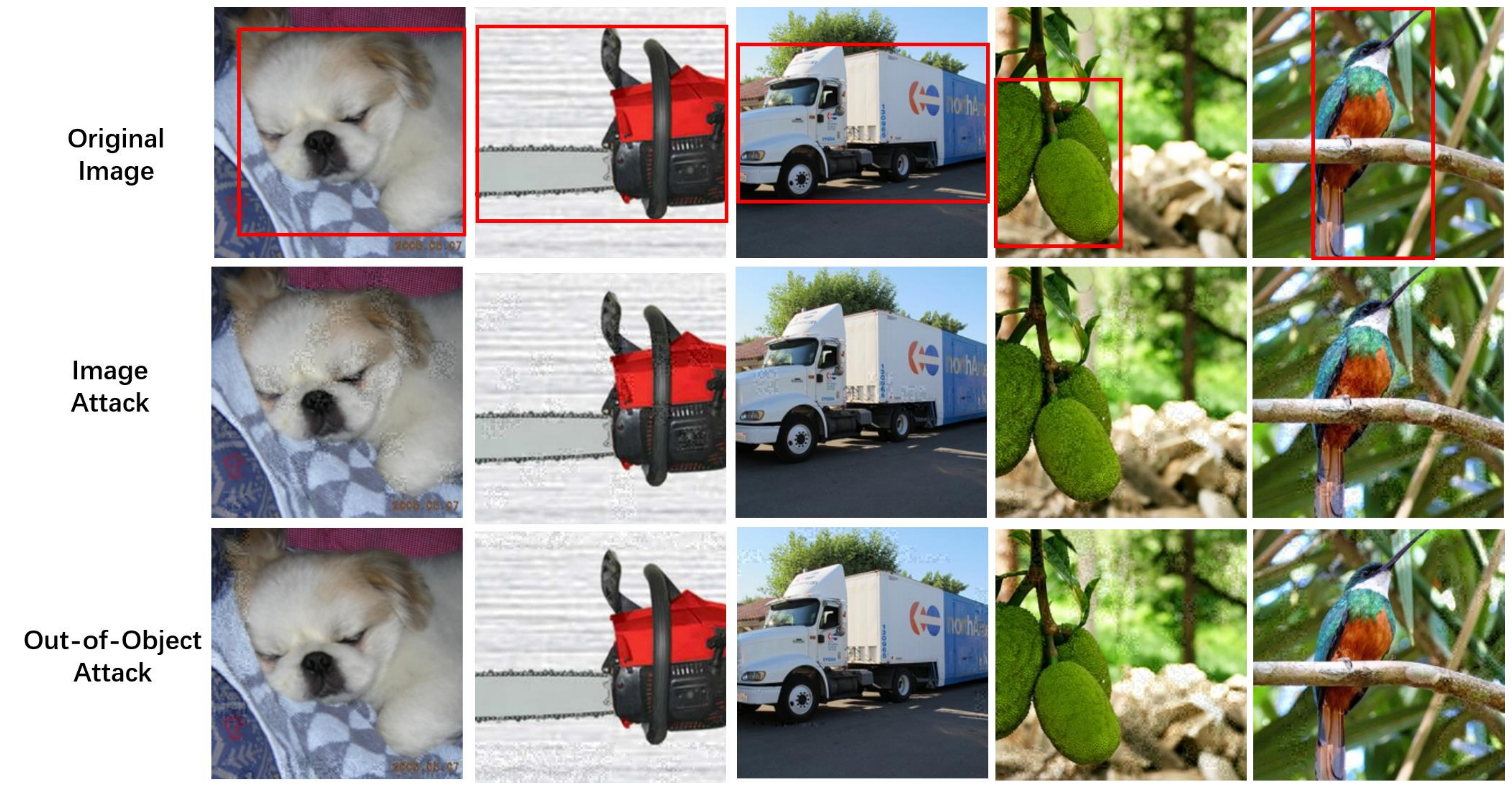}
  \caption{\small{Visualized adversarial examples in out-of-object attack. The red bounding box locates the target object in the original image. In \emph{out-of-object attack}, the adversary is only allowed to perturb pixels that are out of the object bounding box. In \emph{image attack}, the adversary can perturb any pixel in the image.}}
  \label{fig:oattack}
\end{figure*}
\begin{table*}[!t]
		\centering
		\caption{Results of the out-of-object attack on ImageNet when {\upshape $\lambda=10, N=1$} and the perceptual distance metric being {\upshape $1-\text{SSIM}$}. I, R and V represent InceptionV3, ResNet50 and VGG16bn, respectively.}
		\scalebox{.8}[.8]{		
		\begin{tabular}{cccccccccccccccc}\toprule
       Attacked   &\multicolumn{3}{c}{Success} &\multicolumn{3}{c}{\multirow{2}{*}{$1-\text{SSIM}$}} &\multicolumn{3}{c}{\multirow{2}{*}{LPIPS}} &\multicolumn{3}{c}{\multirow{2}{*}{CIEDE2000}} &\multicolumn{3}{c}{Avg.} \cr
        Range &\multicolumn{3}{c}{Rate} &&&&&&&&&&\multicolumn{3}{c}{Queries}\cr \cmidrule{2-4} \cmidrule{5-7} \cmidrule{8-10} \cmidrule{11-13} \cmidrule{14-16}
        &I &R &V  &I &R &V   &I &R &V &I &R &V  &I &R &V \cr\midrule
       Image &\textbf{100\%}  &\textbf{100\%} &\textbf{100\%} &0.078  &0.076 &\textbf{0.072}  &0.096 &0.081 &0.079 &0.692  &\textbf{0.741} &0.699 &\textbf{845} &\textbf{401}  &\textbf{251}\cr
       Out-of-object &90.1\% &93.8\% &94.7\% &\textbf{0.071} &\textbf{0.069}  &0.074 &\textbf{0.081} &\textbf{0.065} &\textbf{0.070} &\textbf{0.678} &0.805 &\textbf{0.687} &4275 &3775 &3104\cr
		\bottomrule
		\end{tabular}			
		}
		\label{tab:range}		
\end{table*}
\paragraph{Sampling frequency.} Sampling frequency decides the size of the sample space of $\delta$. Setting higher frequency means there are more noise values to explore through sampling. In Table \ref{tab:abl}, increasing the sampling frequency from $N=1$ to $N=2$ reduces the perceptual distance to some extent at the cost of lower success rate. On the other hand, further increasing $N$ to $12$ does not essentially reduce the distortion yet lowers the success rate. We set the sampling frequency $N=1$ in the following experiments. Note that the maximum sampling frequency is $N=12$ because the sampling interval in RGB color space (\emph{i.e.}, $255*0.05/N$) would be less than $1$ if $N>12$. See Fig. \ref{fig:vis_samp} for a few adversarial examples from different sampling frequencies.

\paragraph{Noise Distribution.} In the proposed algorithm, we adopt a flexible noise distribution instead of predefining it to be a specific form. Therefore, we conducted ablation studies on assuming the distribution to be a regular form as in NAttack \cite{ld2019}. Specifically, we let the noise distribution be an isometric normal distribution while $\lambda=10$ in the loss function, and perform attacks by estimating mean and variance as  Eq. (10) in \cite{ld2019}. As reported in the tenth row of Table \ref{tab:abl}, under the same experimental setting, it is clear that fixing the noise distribution to be a specific isometric normal distribution degrades the overall performance. We think it is because the distribution that minimizes the perceptual distance is unknown, which might not follow a Guassian distribution or other regular form of distribution. To approximate an unknown distribution, it is better to allow the noise distribution to be a free form as in the proposed approach, and let it be learned by minimizing the perceptual distance.
\subsection{Out-of-Object Attack}
Most existing classification networks \cite{He2015Deep,hu2018senet} are based on Convolutional Neural Network (CNN), which gradually aggregates contextual information in deeper layers. Therefore, it is possible to fool the classifier by just attacking the ``context'', \emph{i.e.}, background that is out of the target object. Attacking just the out-of-object pixels constrains the number and the position of pixels that can be perturbed, which might further reduce the visual distortion caused by the noise. To locate the object in a given image, we exploited the object bounding box provided by ImageNet. An out-of-object mask is then created according to the bounding box such that the model is only allowed to attack pixels that are out of the object, as shown in Fig. \ref{fig:oattack}. In Table \ref{tab:range}, we report results of InceptionV3, ResNet50 and VGG16bn with the maximum queries$=40,000$. The attack is performed on images whose masks are at least $10\%$ large of the image area. The results show that attacking just the out-of-object pixels can also cause misclassification of the object with over $90\%$ success rate. Compared with image attack, the out-of-object attack is more difficult for the adversary in that it requires more number of queries ($4275/3775/3104$) yet has lower success rate ($90.1\%/93.8\%/94.7\%$). On the other hand, the out-of-object attack indeed reduces visual distortion of the adversarial examples on the three networks.
\begin{table}[!h]
		\centering
		\caption{Comparison of the undefended ({\upshape v$3$}) and defended ({\upshape v$3_{\text{adv-ens}4}$}) InceptionV3. The defended InceptionV3 adopts ensemble adversarial training.}
		\scalebox{.7}[.7]{		
		\begin{tabular}{ccccccc}\toprule
    Network &Clean Accuracy &After Attack &$1-\text{SSIM}$ &LPIPS &CIEDE2000 &Avg. Queries \cr\midrule
   v$3$  &75.8\% &0.8\% &0.096 &0.149 &0.862 &531 \cr
    v$3_{\text{adv-ens}4}$ &73.4\% &1.8\% &0.103 &0.154 &0.979 &777 \cr
		\bottomrule
		\end{tabular}			
		}
		\label{tab:defended}	
\end{table}
\subsection{Attack Effectiveness on Defended Network}
\begin{figure*}[!t]
\centering
    \includegraphics[width=0.85\textwidth]{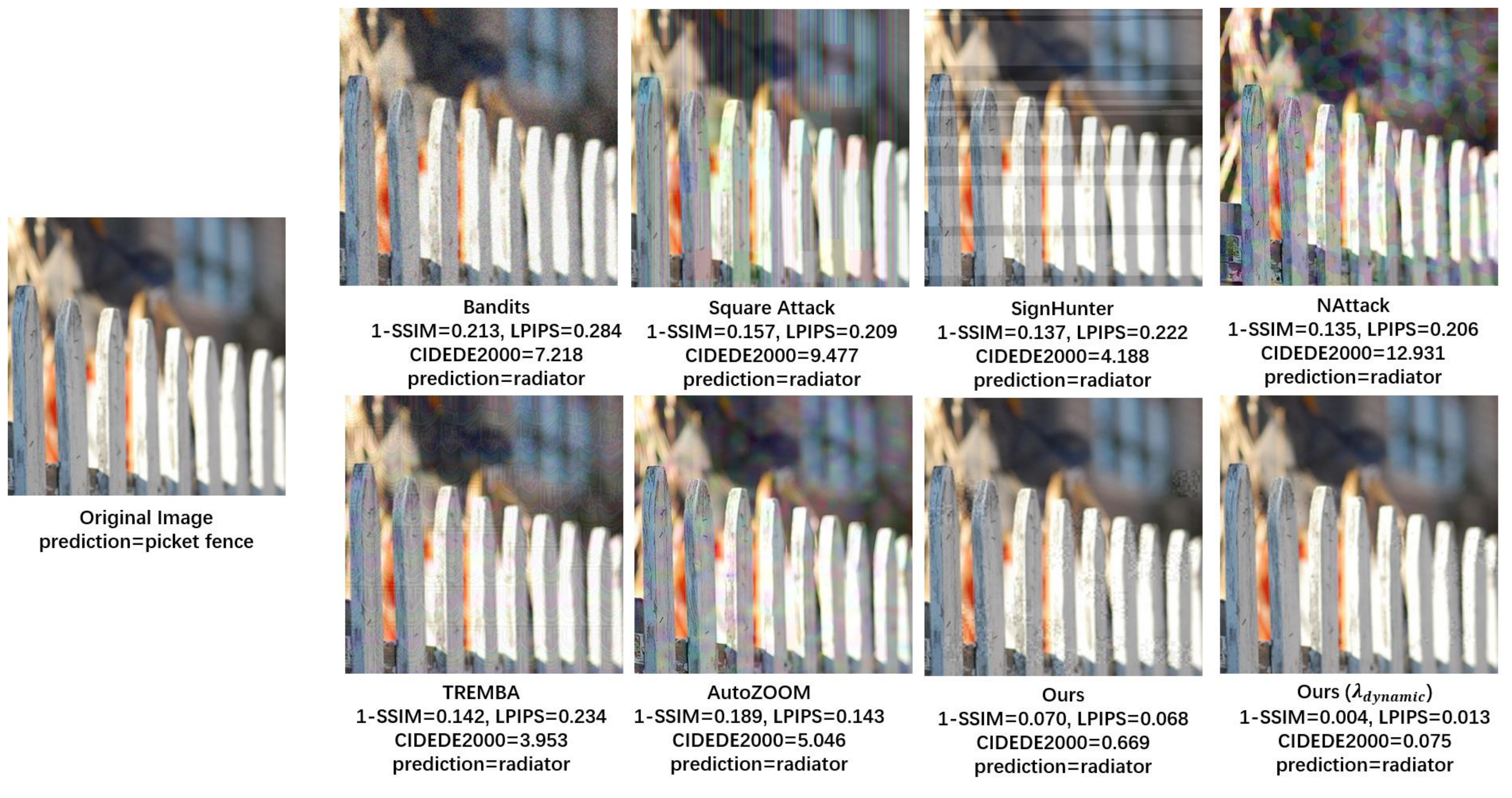}
  \caption{Adversarial examples from different attacks with perceptual distance scores.}
  \label{fig:compare}
\end{figure*}
\begin{table*}[!t]
		\centering
		\caption{Results of different attacks on ImageNet. I, R and V represent InceptionV3, ResNet50 and VGG16bn, respectively.}
		\scalebox{.8}[.8]{		
		\begin{tabular}{cccccccccccccccc}\toprule
        \multirow{3}{*}{Attack}  &\multicolumn{3}{c}{Success} &\multicolumn{3}{c}{\multirow{2}{*}{$1-\text{SSIM}$}} &\multicolumn{3}{c}{\multirow{2}{*}{LPIPS}} &\multicolumn{3}{c}{\multirow{2}{*}{CIEDE2000}} &\multicolumn{3}{c}{Avg.} \cr
        &\multicolumn{3}{c}{Rate} &&&&&&&&&&\multicolumn{3}{c}{Queries}\cr \cmidrule{2-4} \cmidrule{5-7} \cmidrule{8-10} \cmidrule{11-13}  \cmidrule{14-16}
        &I &R &V  &I &R &V   &I &R &V  &I &R &V &I &R &V \cr\midrule
        SignHunter \cite{there20} &98.4\% &- &- &0.157 &- &- &0.117 &- &- &3.837 &- &- &450 &- &-\cr
         NAttack \cite{ld2019} &99.5\% &- &- &0.133 &- &- &0.212 &- &- &5.478 &- &- &524 &- &\cr
         AutoZOOM \cite{TuTC0ZYHC19} &100\%	&-	&-	&0.038	&-	&-	&0.059	&-	&-	&3.33	&-	&-	&1010	&-	&-\cr
        Bandits \cite{ilyas2018prior} &96.5\% &98.8\% &98.2\% &0.343 &0.307 &0.282 &0.201 &0.157 &0.140 &8.383 &8.552 &8.194 &935 &705 &388\cr
        Square Attack \cite{andriushchenko2019square} &99.7\% &\textbf{100\%} &\textbf{100\%} &0.280 &0.279 &0.299 &0.265 &0.243 &0.247 &9.329	&9.425	&9.429 &\textbf{237} &\textbf{62} &\textbf{30} \cr
        TREMBA \cite{r3_5huang2019black} &99.0\%	&\textbf{100\%}	&99.8\%	&0.161	&0.161	&0.160	&0.188	&0.189	&0.187	&4.413	&4.400	&4.421	&-	&-	&-\cr \midrule
        SignHunter-SSIM  &97.6\% &- &- &0.220 &- &- &0.157 &- &- &3.832 &- &- &642 &- &-\cr
         NAttack-SSIM &97.3\% &- &- &0.128 &- &- &0.210 &- &- &5.021 &- &- &666 &- &- \cr
         AutoZOOM-SSIM &\textbf{100\%}	&- &	-	&0.028 &	-	&-	&0.048 &	-	&- &	2.98 &- &-	&2245	&- &-\cr
         Bandits-SSIM &80.0\% &89.3\% &89.7\% &0.333 &0.303 &0.275 &0.200 &0.163 &0.135 &8.838 &8.666 &8.194 &1318 &1020 &793\cr
         Square Attack-SSIM &99.2\% &100\% &100\% &0.260 &0.268 &0.292 &0.256 &0.238 &0.245 &9.301	&9.462	&9.451 &278 &65 &\textbf{30} \cr
         TREMBA-SSIM &98.5\%	&100\%	&99.8\%	&0.160 &0.160	&0.159	 &0.185	&0.186 &0.183 &4.410	&4.396 &4.421 &-	&-	 &-\cr
       Ours &98.7\% &\textbf{100\%} &\textbf{100\%}  &0.075  &0.076 &0.072 &0.094  &0.081 &0.079 &0.692 &0.741 &0.699 &731 &401  &251\cr
       Ours($\lambda_{dynamic}$) &\textbf{100\%} &\textbf{100\%} &\textbf{100\%}  &\textbf{0.016}  &\textbf{0.009} &\textbf{0.006} &\textbf{0.023}  &\textbf{0.009} &\textbf{0.005} &\textbf{0.215} &\textbf{0.204} &\textbf{0.155} &7311 &7678  &7620\cr
		\bottomrule
		\end{tabular}			
		}
		\label{tab:compare}		
\end{table*}
In the above experiments, we show that our black-box model can attack the \emph{undefended} network with high success rate. To evaluate the strength of the proposed attack in \emph{defended} situation, we further attack the InceptionV3 network that adopts ensemble adversarial training (\emph{i.e.}, v$3_{\text{adv-ens}4}$). Following \cite{TramerKPGBM18}, we set $\epsilon=0.0625$ and randomly select $10,000$ images from the ImageNet validation set for test. The maximum number of queries is $10,000$. The performance of the attacked network is reported in Table\ref{tab:defended}, where clean accuracy is the classification accuracy before attack. Note that v$3$ is slightly different from InceptionV3 in Table \ref{tab:abl} in that the pretrained model of v$3$ comes from Tensorflow, which is the same platform of the pretrained model of v$3_{\text{adv-ens}4}$. Compared with undefended network, attacking defended one causes larger visual distortion. However, the proposed attack can still reduce the classification accuracy from $73.4\%$ to $1.8\%$, which demonstrates its effectiveness under defend.
\begin{figure*}[!t]
\centering
    \includegraphics[width=0.9\textwidth]{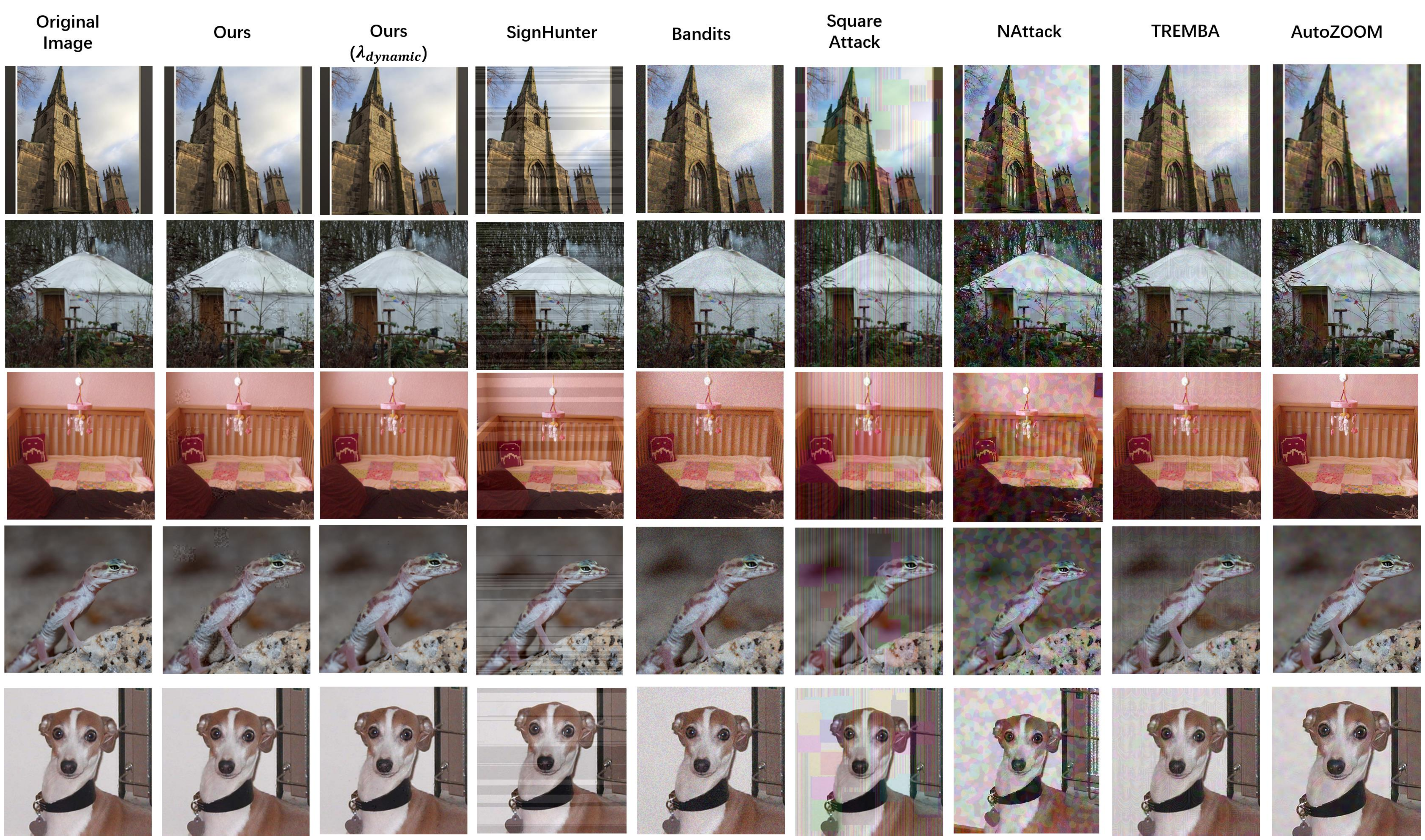}
  \caption{More visualized adversarial examples from different attacks.}
  \label{fig:more}
\end{figure*}
\subsection{Comparison with Other Attacks}
Since our approach addresses improving the visual similarity between the adversarial example and the original image, it might cost more number of queries to construct a less distorted adversarial example. To show that such costs are affordable, we compare our attack to recently proposed black-box attacks: SignHunter \cite{there20}, NAttack \cite{ld2019}, AutoZOOM \cite{TuTC0ZYHC19}, Bandits \cite{ilyas2018prior}, Square Attack \cite{andriushchenko2019square} and TREMBA \cite{r3_5huang2019black}. For fair comparison, in Table \ref{tab:compare}, methods marked with -SSIM and \textbf{Ours} introduce $\lambda \cdot (1-\text{SSIM})$ to the loss function with $\lambda=10$. Note that AutoZOOM performs line search on the choice of $\lambda$, for which we adopt the same strategy and denotes this variant of our method as \textbf{Ours($\lambda_{dynamic}$)}. 
The results of the above methods are reproduced using the official codes provided by the authors. We use the default parameter settings of the corresponding attack, and set the maximum number of queries to be $10,000$. See Table \ref{tab:settings} for the experimental settings of different methods. In Table \ref{tab:compare}, Comparing approaches that use fixed $\lambda$ value (i.e., Signhunter-SSIM, NAttack-SSIM, Bandits-SSIM, Square Attack-SSIM, TREMBA-SSIM, AdvGAN-SSIM and Ours), we can see that the proposed method outperforms other attacks on reducing perceptual distance, while the average number of queries is comparable to Bandits. On the other hand, Ours($\lambda_{dynamic}$) achieves state-of-the-art performance on 1-SSIM, LPIPS and CIEDE2000 when compared with methods that perform line search over $\lambda$ (i.e., AutoZOOM and AutoZOOM-SSIM). In general, except for Signhunter, introducing perceptual distance metric in the objective function helps reduce visual distortion in other attacks. The visualized adversarial examples from different attacks are given in Fig. \ref{fig:compare}, which shows that our model produces less distorted adversarial examples. More examples can be found in Fig. \ref{fig:more}.
\begin{figure}[!t]
\centering
    \includegraphics[width=0.5\textwidth]{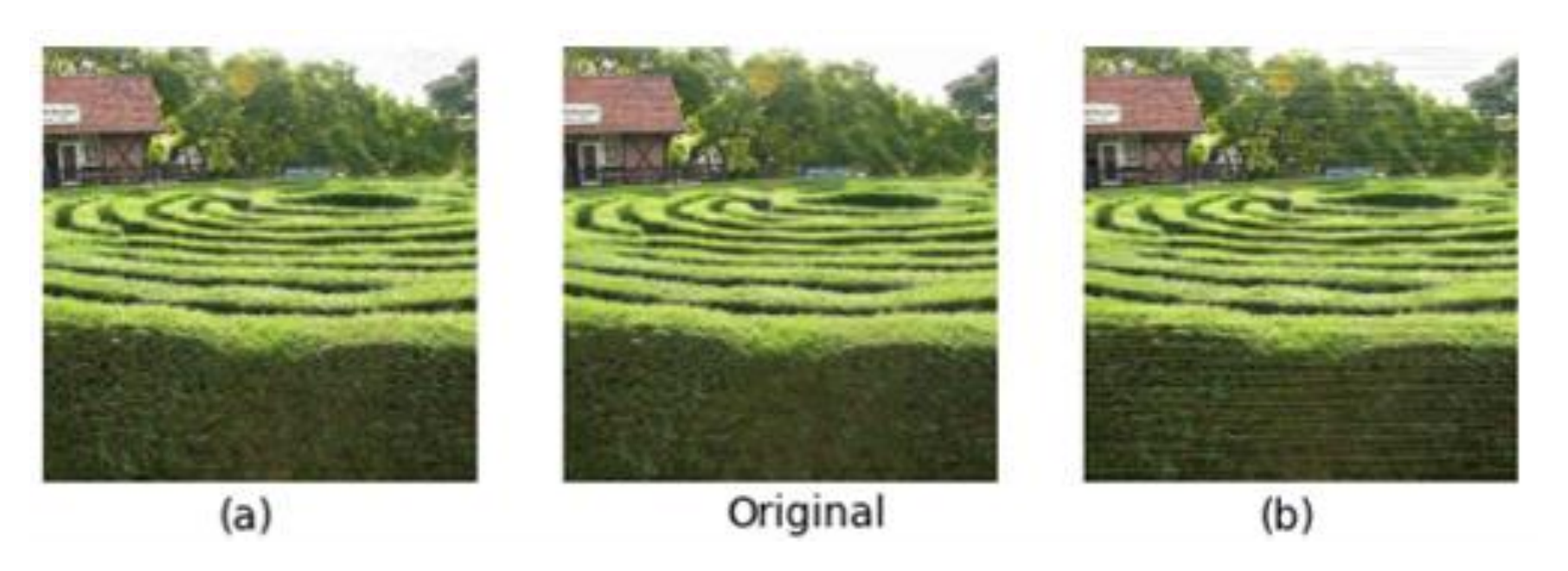}
  \caption{An example of the pictures that we show to the evaluator. One of (a)(b) is produced by our model and the other is from the attacks (excluding ours) in Table \ref{tab:compare}.}
  \label{fig:sub}
\end{figure}

We noticed that adversarial examples from SignHunter have horizontal-stripped noise and Square Attack generates adversarial examples with vertical-stripped noise. Stripped noise is helpful in improving query efficiency since the classification network is quite sensitive to such noise  \cite{andriushchenko2019square}. However, from the perspective of visual distortion, such noise greatly degrades the image quality. The adversarial examples of Bandits are relatively perceptible-friendly, but the perturbation affects most pixels in the image, which causes visually ``noisy'' effects, especially in a monocolor background. The noise maps from NAttack and AutoZOOM appear to be regular color patches all over the image due to their large tiling size in the methods.
\begin{table}[!t]
		\centering
		\caption{Experimental settings.}
		\scalebox{.9}[.9]{		
		\begin{tabular}{ccc}\toprule
   Method &$\lambda$ &Max. Iterations\cr\midrule
  Signhunter-SSIM	&10	&10,000 \cr
  NAttack-SSIM	&10	&10,000 \cr
  AutoZOOM-SSIM	 &dynamic, $\lambda \in [0,1000]$	&10,000 \cr
  Bandits-SSIM	&10	&10,000 \cr
  Square Attack-SSIM	&10	&10,000 \cr
  TREMBA-SSIM	&10	&- \cr
  Ours	&10	&10,000 \cr
  Ours($\lambda_{dynamic}$)	&dynamic, $\lambda \in [0,1000]$	 &10,000 \cr
		\bottomrule
		\end{tabular}			
		}
		\label{tab:settings}	
\end{table}

We also conducted subjective studies for further validation. Specifically, we randomly chose two adversarial examples, where one is generated by our approach (Ours($\lambda_{dynamic}$)) and the other is given by the attacks (excluding ours) in Table \ref{tab:compare}. We show each human evaluator the two adversarial examples, and ask him/her which one is less distorted compared with the original image. Figure \ref{fig:sub} gives an picture that we show to the evaluator. Note that the order of the two adversarial examples in the triplet is randomly permuted. We asked 10 human evaluators in total, each made judgements over $100$ triplets of images. As a result, adversarial examples generated by our method are thought to have less noticeable noise $82.1\%$ of the time, while $10.0\%$ of the time the evaluators think both examples are distorted at the same level. Therefore, the subjective results further prove that the proposed method effectively reduces visual distortion in adversarial examples.
\begin{table*}[!t]
		\centering
		\caption{Results of other {\upshape $l_p$} attacks on ResNet50 when {\upshape $\lambda=10$}. The raw {\upshape $l_0$} and {\upshape $l_1$} scores have much higher order of magnitude compared with other metrics, and thus the normalized scores of {\upshape $l_0$} and {\upshape $l_1$} distances are reported. }
		\scalebox{.8}[.8]{		
		\begin{tabular}{cccccccccc}\toprule
     Distance Metric &Sampling Frequency &Success Rate &$1-\text{SSIM}$ &LPIPS &CIEDE2000 &$l_0$ &$l_1$ &$l_2$ &Avg. Queries \cr \midrule
     \multirow{3}{*}{$l_0$} &1 &\textbf{99.5\%} &0.077 &0.083 &0.795 &\textbf{0.133} &0.130 &6.75 &536\cr 
     &2 &99.2\% &0.065 &0.069 &\textbf{0.768} &0.159 &0.118 &5.88 &679 \cr
     &5 &97.9\% &\textbf{0.058} &\textbf{0.065} &0.789 &0.177 &\textbf{0.118} &\textbf{5.19} &960 \cr\midrule
       \multirow{3}{*}{$l_1$} &1 &\textbf{99.5\%}  &0.077 &0.083 &0.795 &\textbf{0.133} &0.130 &6.75 &536\cr 
       &2 &\textbf{99.5\%} &0.070  &0.076 &0.773 &0.176 &0.130 &6.14 &658\cr 
       &5 &99.2\% &0.066 &0.070 &0.768 &0.218 &0.129 &5.74 &800 \cr \midrule
        \multirow{3}{*}{$l_2$}  &1 &\textbf{99.5\%} &0.110 &0.112 &0.829 &0.215 &0.211 &8.21&\textbf{392}\cr
        &2 &\textbf{99.5\%} &0.092 &0.100 &0.803 &0.259 &0.191 &7.44 &431 \cr
        &5 &\textbf{99.5\%} &0.087 &0.094 &0.792 &0.312&0.185 &6.89 &579\cr 
		\bottomrule
		\end{tabular}			
		}
		\label{tab:lp}		
\end{table*}
\subsection{Other $l_p$ Attacks}
\label{sec:lp}
Although our method in this paper is based on $l_\infty$ attack,  the perceptual distance metric $d$ in the loss function can be replaced by other $l_p$ ($p=0,1,2$) distance. We did not discuss it in the above experiments because these $l_p$ distance metrics are less accurate in measuring the \emph{perceptual} distance between images compared to the specifically designed metrics, such as well-established $1-\text{SSIM}$ and LPIPS. Nevertheless, we still present the results of other $l_p$ ($p=0,1,2$) attacks in Table \ref{tab:lp}, where the $l_p$ distance is normalized to $[0,1]$ in the loss function. Specifically, $d(x,x+\delta)=\frac{{{l_p}(x,x + \delta )}}{{{{\max }_\delta }({l_p}(x,x + \delta ))}}$, where ${l_p}(x,x + \delta)$ is the $l_p$ distance between the original image $x$ and the perturbed image $x+\delta$. As in the paper, we set $\lambda=10,\epsilon=0.05$ and the maximum number of queries being $10,000$. We find that the raw $l_0$ and $l_1$ scores have much higher order of magnitude compared with other metrics, and thus the normalized scores of $l_0$ and $l_1$ distances are reported in Table \ref{tab:lp}. Note that when the sampling frequency $N=1$, $l_0$ distance is equivalent to $l_1$ distance in that
\begin{equation}
\begin{aligned}
\frac{{{l_1}(x,x + \delta )}}{{{{\max }_\delta }({l_1}(x,x + \delta ))}}& = \frac{{mc \cdot \epsilon }}{{WHc \cdot \epsilon }} \\
&= \frac{m}{{WH}} \\
&= \frac{{{l_0}(x,x + \delta )}}{{{{\max }_\delta }({l_0}(x,x + \delta ))}}
\end{aligned}
\end{equation}
where $m$ is the number of perturbed pixels. $W,H$ and $c$ are the width, height and number of channels of a given image, respectively. Table \ref{tab:lp} shows that optimizing $l_0$ distance gives better performance on both the perceptual distance metrics and the $l_p$ distance metrics.
\subsection{Conclusion}
We introduce a novel black-box attack based on the induced visual distortion in the adversarial example. The quantified visual distortion, which measures the perceptual distance between the adversarial example and the original image, is introduced in our loss where the gradient of the corresponding non-differentiable loss function is approximated by sampling from a learned noise distribution. The proposed attack can achieve a trade-off between visual distortion and query efficiency by introducing the weighted perceptual distance metric in addition to the original loss. The experiments demonstrate the effectiveness of our attack on ImageNet as our model achieves much lower distortion when compared to existing attacks. In addition, it is shown that our attack is valid even when it's only allowed to perturb pixels that are out of the target object in a given image.



\normalsize
\bibliography{myBib}


\end{document}